%% file: paclic2023.tex
\pdfoutput=1

\documentclass[11pt]{article}

\usepackage[]{acl2023}

\usepackage{times}
\usepackage{latexsym}
\usepackage{paralist}
\usepackage[compact]{titlesec}

\usepackage[T1]{fontenc}

\usepackage[utf8]{inputenc}

\usepackage{microtype}
\usepackage{subfig}
\usepackage{wrapfig}

\usepackage{inconsolata}
\usepackage{float}
\usepackage{mathtools,amssymb,lipsum, nccmath, bm}
\usepackage{color, colortbl}
\NewDocumentCommand{\up}{som}{
  \IfBooleanTF{#1}
    {\upext{#3}}
    {#3\IfNoValueTF{#2}{\mathord}{#2}\uparrow}%
}
\renewcommand{\and}{\\}
\title{Improving Cross-Domain Hate Speech Generalizability with Emotion Knowledge}

\author{Shi Yin Hong \and Susan Gauch\\
  Department of Electrical Engineering and Computer Science \\
  University of Arkansas, Fayetteville, AR, USA \\
  \texttt{\{syhong, sgauch\}@uark.edu }
  }

\begin{document}
\maketitle
\begin{abstract} 
Reliable automatic hate speech (HS) detection systems must adapt to the in-flow of diverse new data to curtail hate speech. However, hate speech detection systems commonly lack generalizability in identifying hate speech dissimilar to data used in training, impeding their robustness in real-world deployments. In this work, we propose a hate speech generalization framework that leverages emotion knowledge in a multitask architecture to improve the generalizability of hate speech detection in a cross-domain setting. We investigate emotion corpora with varying emotion categorical scopes to determine the best corpus scope for supplying emotion knowledge to foster generalized hate speech detection. We further assess the relationship between using pretrained Transformers models adapted for hate speech and its effect on our emotion-enriched hate speech generalization model. We perform extensive experiments on six publicly available datasets sourced from different online domains and show that our emotion-enriched HS detection generalization method demonstrates consistent generalization improvement in cross-domain evaluation, increasing generalization performance up to 18.1\% and average cross-domain performance up to 8.5\%, according to the F1 measure\footnote{Code and resources are available at \url{https://github.com/sy-hong/ek-hs-generalizability}}. 
\end{abstract}

\section{Introduction}
Hate speech (HS) possesses justified regulatory grounds since it inflicts harm toward a targeted individual or group based on perceived characteristics \citep{gelber2021differentiating}. The social obstruction imposed by the pervasive online HS thus ignites counteraction from the natural language processing (NLP) community to create machine learning-based systems to automate the HS identification process \citep{poletto2021resources, info13060273}. Despite efforts dedicated to the goal in the past decade, HS detection remains a challenging task to conquer \citep{fortuna-directions, wiegand21}. Notably, the lack of generalizability is a prevalent issue with current HS models \citep{yin2021towards}.

\input{figtab/data}

HS models that suffer from generalizability show a discrepancy in their performance across HS datasets \citep{wiegand2019detection}. Such models are competitive in detecting HS on the data from the same source as the data they are trained with but show a significant performance gap when detecting HS from varied HS sources. The mainstream approach addressing the issue utilizes knowledge from the HS domain to improve HS generalization. Observations on the semantic distribution of the data (e.g., implicit or explicit HS) serve as the basis for counteractive augmentation, synthetic data generative, and sampling techniques to bridge linguistic gaps observed in HS \citep{harald22, arango2022hate, wullach21, ludwig22}. Furthermore, datasets may be re-annotated, combined, or created to meet the generalization task by topics \citep{yoder22, nejadgholi22, toraman22}. 

In real-world applications, however, it is unrealistic to conduct retrospective HS analysis with the constant inflow of new and changing data. Further, HS is the byproduct of the evolving social context, culture, and linguistic interpretation \citep{hilte2023haters}, which elevates the challenge of using static criteria in assessing hate speech. HS models that cannot demonstrate robust generalizability cannot reliably carry out their high-stake social responsibility in safeguarding vulnerable groups from the multifarious HS reflected in diverse online platforms. The lack of generalizability of HS models can even unintentionally exacerbate the proliferation of online HS by allowing out-of-domain hateful speech to evade its consequences while curtailing free speech when sanctioning unhateful speech \citep{just}. 

 In this work, we utilize emotion knowledge to support the generalization of HS detection. We find that utilizing emotion knowledge in addressing HS, which exhibits a greater relative conceptual variability, helps to mitigate the variance of HS language that challenges HS generalization. Specifically, we adopt the GoEmotions dataset \citep{goemotions} to provide emotion information and investigate the effect of leveraging two variants of emotion corpora -- the dataset's original release with 28 emotions and its Ekman emotion corpus \citep{ekman1971universals} equivalent -- in improving the HS detection's generalizability in a multitask framework. We utilize BERT \citep{devlin} and fBERT \citep{sarkar} as the base Transformers models to evaluate their effectiveness in enhancing emotion-driven HS generalizability given their varying pre-trained corpora relatedness to HS. We assess the proposed model's performance in improving the generalizability of six popular benchmark datasets from different domains with the cross-dataset evaluation method. Our emotion-enriched HS detection generalization method demonstrates consistent cross-domain generalization binary F1 performance, increasing generalization performance up to 18.1\% and average cross-domain up to 8.5\%. Our main contributions are summarized below:
\begin{compactitem}
    \item We propose an emotion-integrated multitask HS generalization framework that utilizes emotion knowledge to strengthen cross-domain HS generalization.
    \item We study how the categorical scope of the emotion corpora -- the 28-class GoEmotions \citep{goemotions} corpus and the six-class Ekman emotion \citep{ekman1971universals} corpus -- affects the generalizability of HS detection with our method.
    \item We evaluate the effect of the adopted Transformers base models', BERT and fBERT, varying adaptiveness to the HS domain on our cross-domain HS generalization framework.
    \item We perform extensive evaluations in cross-domain settings on six publicly available benchmark datasets with varied HS forms to show our model improves cross-domain HS generalization.
\end{compactitem}

\section{Related Works}
\subsection{Generalization of HS Detection}
In studies of HS detection's generalization, some works aim to identify sources behind the lack of performance generalization. \citet{fortuna20} analyze the homogeneity of applied categories in popular public HS datasets and empirically support the lack of compatibility among the cross-dataset performance. They suggest an underlying reason is the lack of consensus on HS's subjective definitional concept, leading to varied criteria for HS categorization. \citet{fortuna2021well} reason that the low generalization of HS detectors roots to the imbalance of explicit and implicit distributions of HS across datasets and encourage HS dataset creators to identify precise categorization (e.g., sexism, racism) to endow levels of granularity in HS. \citet{arango2022hate} argue for more transparency behind the user distribution of existing HS datasets to prevent spurious high performance of HS classifiers overfitted to limited users in data production. They propose to improve cross-dataset generalization by adopting countering sampling techniques addressing user overfitting.

\input{figtab/emc}

The dominant method in addressing the generalization of HS detection considers analyzing underlying semantic and topical traits of HS datasets. \citet{bourgeade23} sample from six HS corpora and present a re-annotated dataset version based on topic-generic and topic-specific levels. They find that adopting a mixture of topic-generic and topic-specific tweets in the model fine-tuning step enhances the generalization of HS classifiers. \citet{nejadgholi22} show the weakness of HS classifiers at generalizing implicit racism from topic-centric HS datasets and propose a model based on concept activation vector to improve the interpretability of the model in performing generalization. \citet{ludwig22} adopt unsupervised domain adaptation to improve HS models’ ability to perform generalization across HS targeting toward a subset of categorized target groups encompassed by the HateXExplain dataset \citep{mathew2021hatexplain}. \citet{wullach21} adopt a GPT-based language model to generate synthetic HS via sequence generation using existing HS datasets as an augmentation technique to improve the quality of HS generalization.

\subsection{Multitask HS Detection With Emotions}
Multitask learning is a training methodology that has recently gained popularity in NLP due to its power to integrate knowledge from related tasks in modeling a target task \citep{zhang23, turcan21}. In the context of this study, emotion classification is the auxiliary task modeled jointly with the main task of binary hate speech classification to improve hate speech detection generalization leveraging the integrated emotion knowledge. 
Present works that consider emotion features in related studies focus more on the abusive language detection domain without considering the model's generalizability. \citet{raja2020} investigate abusive language detection by incorporating emotion detection into MLP-based and BiLSTM-based networks with a hard-sharing multitask framework. \citet{sam2019} employ pre-trained DeepEmoji to assign textual data with relevant emotions in capturing offensive language. \citet{halat2022multi} and \citet{plaza2022integrating} incorporated sentiment information in addressing HS domain-related tasks, offensive and abusive language detections, with emotion information. In the HS domain, \citet{chiril2022emotionally:23} adopt emotion lexicons such as SenticNet and EmoSenticNet to detect hate speech with multi-targets from topic-generic datasets and conclude that the utilization of affective knowledge enhances hate speech detection categorized by targets and topics. \citet{mnassri} perform hate and offensive language detection using emotion information in a multitask setting involving cross-lingual settings.  

Diverging from previous works, we utilize emotion knowledge with varying categorical scopes to uplift cross-domain generalizability. We further examine Transformers models' relative domain adaptability to the HS in cross-domain generalizability.

\section{Methodology}
\subsection{Experimental Datasets}
We adopted the six datasets and the processing procedures used by \citet{harald22}, a recent hate speech generalization work, to assess our model's ability to improve hate speech detection generalizability in a cross-domain setting. These datasets are: Founta \citep{founta2018large}, Kaggle \citep{kaggle}, Kumar \citep{kumar}, Waseem and Hovy (W\&H) \citep{waseem}, Offensive Reddit \citep{qian}, and Razavi \citep{razavi}. For larger datasets such as Founta, Kaggle, and Kumar, positive and negative HS samples were randomly sampled for 5,000 entries, totaling 10,000 samples per dataset. Negative samples in smaller datasets such as Razavi and W\&H were downsampled to the sizes of positive samples, 482 and 795, respectively, to control the source of variance in generalization study \citep{swamy19}. For the Offensive Reddit dataset, the 3,230 positive samples were balanced with 3,230 negative samples from \citet{subred}. User mentions, hashtags, URLs, and emojis were removed for text preprocessing. Table 1 shows an overview of the datasets and their respective domains used in our cross-domain HS generalizability study.

\begin{figure}
\subfloat{
\includegraphics[width=\columnwidth]{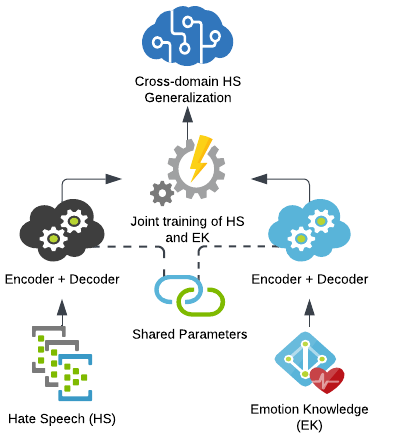}
}
\caption{Abstract architecture for the emotion knowledge-enriched HS generalization framework.}
\label{fig:arch}
\end{figure}

We use two variants of the GoEmotions dataset \citep{goemotions} to supply the emotion information. The first variant is the original version of the dataset with 27 emotions plus the neutral class, which we would refer to as $GE_{go}$ in the following sections. $GE_{go}$ serves in our study by providing fine-grained categorical emotion information in assessing the generalizability of our emotion-enriched hate speech detection method. To contrast the detailed emotion categories supplied by $GE_{go}$, we convert the 28 emotion classes in $GE_{go}$ into its Ekman equivalent with six emotions as detailed in \citet{goemotions}. We refer to the GoEmotions dataset in the form of the traditional six-class emotion corpus as $GE_{ek}$. Table 2 provides the emotion categorical conversion between $GE_{go}$ and $GE_{ek}$.

\input{figtab/exp1}
\subsection{Emotion-enriched Hate Speech Generalization}

We implement a multitask architecture as illustrated in Figure 1 given a set of disjoint tasks $\Omega = \{\Omega_{hs}, \Omega_{e}\}$, where $\Omega_{hs}$ denotes the main task of hate speech detection and $\Omega_{e}$ denotes the auxiliary task of emotion analysis. We let $\Omega_{hs}$ and $ \Omega_{e}$ share the same Transformers-based encoding layers to promote direct knowledge transfer. This hard parameter-sharing choice prevents overfitting and overparameterization \citep{ruder2017overview} \citep{corr19}. The respective dataset $D_\Omega$ for each task is $D_{hs}$ for $\Omega_{hs}$ and $D_{e}$ for $\Omega_{e}$, where $D_{\Omega} = \{(\chi_{\Omega}^{(i)}, y_{\Omega}^{(i)})\}|_{i=1}^{N}$. The input and target output are denoted by $\chi_{\Omega}^{(i)}$ and $y_{\Omega}^{(i)}$, and  $N$ is the total data entry for each task. 

Given $D_{\Omega}$, we first tokenize $\omega \in D_{\Omega}$ into their subword representations, $s_{\omega}$. We obtain the embedding vector $v(s_{\omega})$ via an embedding layer that transforms $[e_{1}(s_{\omega}), e_{2}(s_{\omega}), ..., e_{d}(s_{\omega})]^T$ into their vectorial representation, where $d$ is the dimension of the embedding space and $e_{i}(s_{\omega})$ denotes the $i^{th}$ element of the embedding vector. We acquire the hidden state $h_{\Omega}^{(i)}$ as follows:
\begin{equation}
h_{\Omega}^{(i)} = Encoder(v(s_{\omega})^{(i)}, \xi_{\Omega}^E)
\end{equation}
where $\xi_{\Omega}^E$ denotes the parameters for the encoder. We use BERT \citep{devlin} and fBERT \citep{sarkar} for encoding, where BERT is trained on a general corpus, and fBERT is trained on an offensive language corpus based on the OLID dataset \citep{zampieri}. We promote the joint knowledge exchange between $\Omega_{hs}$ and $\Omega_e$ as their parameters are shared in the encoder unit. In the process, $\Omega_e$ functions as a regularizer, introducing an inductive bias as the two tasks share more general representations that make the model favors prediction $\hat y_{hs}^{(i)}$ that explains both tasks well.

\input{figtab/exp2}

Both tasks $\Omega_{hs}$ and $\Omega_{e}$ continue to share parameters at the decoding stage but are independent with separate MLP layers for each task. For each task, the predicted output $\hat y_{\Omega}^{(i)}$ is obtained as follows:
\begin{equation}
\hat y_{\Omega}^{(i)} = Decoder(h_{\Omega}^{(i)}, \xi_{\Omega}^D)
\end{equation}
where $\xi_{\Omega}^D$ denotes the parameters of the decoder for $\Omega$. 

For single-class prediction when modeling $\Omega_{hs}$ and $\Omega_e$ using the $GE_{ek}$ corpus, we minimize the negative log-likelihood (NLL) loss: 

\begin{equation}
\begin{split}
    L_{NLL}(\Omega) = & -\sum_{i=1}^{n}(y_{\Omega}^{(i)} log(\hat{y}^{(i)}_{\Omega}) + \\ 
                      & (1-y_{\Omega}^{(i)})log(1-\hat{y}^{(i)}_{\Omega}))
\end{split}
\end{equation}

where $y_{\Omega}^{(i)}$ denotes the ground truth label, and $\hat{y}^{(i)}_{\Omega}$ denotes the predicted label for $\Omega$.
For making predictions using the multi-labeled $GE_{go}$ corpus, we apply the binary cross-entropy (BCE) loss:

\begin{equation}
\begin{split}
 L_{BCE}(\Omega) = & -\frac{1}{N}\sum_{i=1}^N y_{\Omega}^{(i)} log(p(y_{\Omega}^{(i)})) + \\
 & (1-y_{\Omega}^{(i)})log(1-p(y_{\Omega}^{(i)}))
\end{split}
\end{equation}

where $N$ is the training entry count, $y_{\Omega}^{(i)}$ denotes the ground truth label and $p(y_{\Omega}^{(i)})$ denotes the prediction probability for true positive prediction for $\Omega$.

\section{Cross-Domain Generalization}
\subsection{Experimental Setup and Implementation Details}
We compare the performance of our approach with the uncased base version of BERT and fBERT fine-tuned only with HS datasets as baselines. For our emotion-integrated HS generalization model, we adopt BERT and fBERT as the base models as shown in Table 3 and Table 4, respectively. Emotion-integrated models are noted with their respective emotion corpus, $+ GE_{go}$ or $+ GE_{ek}$. We perform training on one dataset and evaluation on all datasets' separate testing sets. This includes in-dataset evaluation as we perform training and testing on the same dataset, which also assesses the in-domain generalizability. We assess cross-domain generalizability performance between different training and test sets not from the same domain. We further analyze the overall generalizability performance by providing the average cross-domain binary F1 for individual experiments as shown in the last column (CD Avg) of Table 3 and Table 4. 

All models were implemented using PyTorch \citep{paszke2019pytorch}, and all experiments were conducted on NVIDIA Quadro RTX 4000. We trained all models with 5 epochs with early stopping as we often observed that the best validation performance is obtained in the first three epochs. We employed a batch size of 8 and an Adam optimizer with 1E-4 as the learning rate. The average binary F1 performance of three separate trails using seeds \{0, 1, 3\} are reported. The best in-domain and cross-domain average scores are in bold. Results that show improvement from baselines are highlighted in gradients of purple for in-domain settings and blue for cross-domain settings based on their relative strength of improvement.

\subsection{Results}
Table 3 shows the performance of our evaluation using BERT as the base model. From the baseline model, we observe a general decline in performance when models are evaluated in a cross-domain setting compared to an in-domain setting, which supports our motivation to improve cross-domain generalizability. The disparity in performance is the greatest for the Offensive Reddit, W\&H, and Kaggle datasets, which show a cross-domain performance decline of 33.7\% $(\frac{0.617-0.931}{0.931} \times 100)$, 22.4\%, and 22.2\%, respectively, when average cross-domain performance is compared with the respective in-domain performance for each evaluation dataset. 

When we include emotion knowledge using the original GoEmotions dataset with its $GE_{go}$ corpus, the performance of in-domain and cross-domain generalization improves. The greatest in-domain improvement ($\up{}$4.5\% = $\frac{\frac{0.597+0.899}{2} - \frac{0.535+0.896}{2}}{\frac{0.535+0.896}{2}} \times 100$) is with the W\&H dataset. Adding the emotion knowledge from the $GE_{go}$ corpus leads to an average increase in cross-domain performance from 2.9\% ($\frac{0.721-0.701}{0.701} \times 100$) to 6.6\% ($\frac{0.709-0.665}{0.665} \times 100$) observed in the Kaggle and Kumar datasets, respectively. Experiments that exhibit higher average out-domain improvement used Kumar ($\up{}$6.6\%), Offensive Reddit ($\up{}$6.6\%), and W\&H ($\up{}$4.9\%) as training sets. The greatest generalizability uplifts for individual cross-domain experiments are shown in Founta $\xrightarrow{}$ Kumar (i.e., Founta dataset is the training set for the model that generalizes on the testing set from the Kumar dataset), Kumar $\xrightarrow{}$ Kaggle, and Razavi $\xrightarrow{}$ Kumar experiments, resulting in generalizability enhancement of 18.1\%, 14.5\%, and 12.2\%, respectively.

When we adopt the GoEmotions dataset based on the $GE_{ek}$ corpus to supply emotion knowledge in our HS generalization model, both in-domain and cross-domain performances also show improvements. The in-domain performance increases up to 4.9\% with the maximum uplifts corresponding to the W\&H datasets. In cross-domain experiments, the average increase in binary F1 ranges from 2.1\% to 8.5\% with the least and best cross-domain performance average corresponding to experimental settings where Founta and Kumar datasets are used as training sets. Integrating emotion knowledge via the $GE_{ek}$ corpus also shows competitive cross-domain generalizability when W\&H ($\up{}$6.4\%) and Kaggle ($\up{}$4.8\%) datasets. The greatest generalizability uplifts for individual cross-domain experiments are shown in Razavi $\xrightarrow{}$ Kumar ($\up{}$17.1\%), Kumar $\xrightarrow{}$ Kaggle ($\up{}$15.7\%), and Kaggle $\xrightarrow{}$ W\&H ($\up{}$10.9\%) experiments. Experiments Razavi $\xrightarrow{}$ Kumar and Kumar $\xrightarrow{}$ Kaggle are also the top-performance individual cross-domain experiments using the $GE_{go}$ corpus.

Table 4 shows the performance of our evaluation when we used fBERT as the base model. We also observe a general decline in performance when models are evaluated in cross-domain settings compared to in-domain settings with the baseline model's performance. The same three datasets when using the BERT as the baseline show the greatest difference in in-domain and cross-domain, resulting in a performance decline of 33.9\%, 19.6\%, and 16.0\% for the Offensive Reddit, Kaggle, and W\&H datasets, respectively. 

\begin{figure}
\subfloat{\includegraphics[width=\columnwidth]{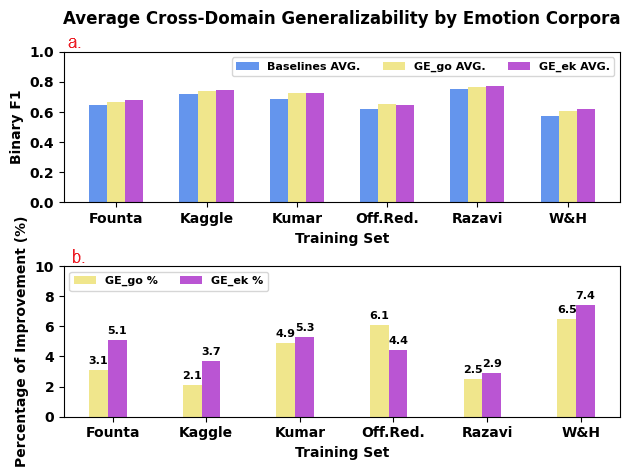}}
\caption{\textbf{a.} Average binary F1 cross-domain generalizability performance of the baseline average, $GE_{go}$-based model average, and $GE_{ek}$-based model average. \textbf{b.} Percentage uplifts compared to the baseline average of the $GE_{go}$-based model average and and $GE_{ek}$-based model average.}
\label{fig:v1}%
\end{figure}

Using the $GE_{go}$ corpus to induce emotion knowledge in our model, we consistently observe cross-domain generalizability enhancement but not always with in-domain experiments. For the Offensive Reddit and Razavi datasets, the in-domain performance decreases by 0.003 in binary F1. 
The greatest in-domain improvement ($\up{}$3.8\%) is observed from the W\&H dataset. The generalizability improvement in average cross-domain performance ranges from 1.2\% to 8.1\%. The minimum and maximum average cross-domain performances correspond to Razavi and W\&H datasets, respectively. Emotion knowledge provided by the $GE_{go}$ corpus also distinctly helps average cross-domain generalizability performance when Offensive Reddit ($\up{}$5.7\%)
and Kumar ($\up{}$3.3\%) are used as the training sets. The individual cross-domain generalizbility enhancement is most pronounced with the W\&H $\xrightarrow{}$ Razavi ($\up{}$15.4\%), W\&H $\xrightarrow{}$ Kumar ($\up{}$13.6\%), and W\&H $\xrightarrow{}$ Founta ($\up{}$9.1\%).

When we adopt the $GE_{ek}$ corpus with fBERT as the base model in our emotion-enriched framework, we observe performance improvement in both in-domain and cross-domain settings. The greatest performance in-domain uplift is 3.5\% with the Founta dataset. The average increase in binary F1 ranges from 0.6\% to 8.3\% in cross-domain settings with the least and best generalizability corresponding to the Razavi and W\&H training sets. 
Strong average cross-domain generalizability enhancement also manifests in experiments where Founta ($\up{}$5.9\%) and Offensive Reddit ($\up{}$4.0\%) datasets are the training sets. The greatest generalizability uplifts for individual cross-domain experiments are shown in  W\&H $\xrightarrow{}$ Razavi ($\up{}$14.6\%), W\&H $\xrightarrow{}$ Kumar ($\up{}$14.1\%), and Founta $\xrightarrow{}$ Razavi ($\up{}$10.3\%). Experiments  W\&H $\xrightarrow{}$ Razavi and W\&H $\xrightarrow{}$ Kumar are also the top-performance individual cross-domain experiments using the $GE_{go}$ corpus.

\begin{figure}
\subfloat{\includegraphics[width=\columnwidth]{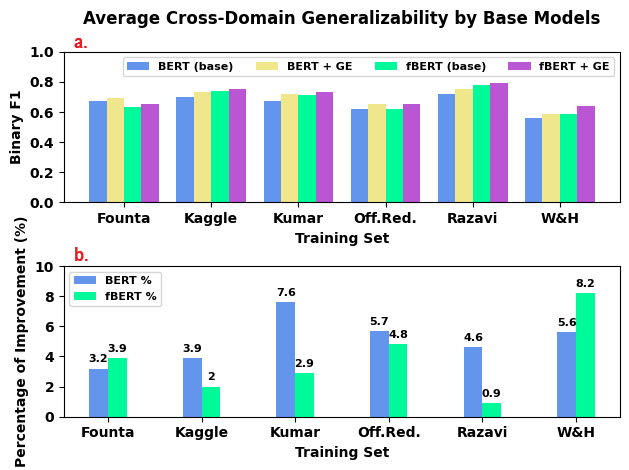} }%
\caption{\textbf{a.} Average binary F1 cross-domain generalizability performance of the baseline average and emotion-enriched model average of the BERT-based and fBERT-based models. \textbf{b.} Percentage uplifts compared to the baseline average of the BERT-based model average and fBERT-based model average.}
\label{fig:v2}
\end{figure}

\section{Analysis and Discussions}
In this section, we study the factors in our framework that affect cross-domain generalizability based on the experimental results. 

\subsection{Emotion Corpora and Cross-Domain HS Generalizability}
\textbf{Observation 1:} \textit{Adopting emotion corpora with fewer categorical scopes, such as the $GE_{ek}$, generally results in more consistent cross-domain HS generalizability improvement.}

To analyze the relationship between the categorical scope of emotion corpus and cross-domain generalization, we visualize the average performance of the cross-domain generalization of two variants of our emotion-enriched models eliminating the distinction of the employed base models. We take the average performance of the cross-domain generalizability of BERT-based and fBERT-based baselines and emotion-enriched models separated by the use of different emotion corpora. The average performance in binary F1 of the baseline, $GE_{go}$-based model, and $GE_{ek}$-based model are shown in Figure 2a. Figure 2b shows the percentage of cross-domain improvement relative to the baseline average.

We observe that the emotion-enriched model where the emotion knowledge is introduced by the $GE_{ek}$ exhibits better performance in all cases except the experiment where the Offensive Reddit dataset is used as the training set. The highest generalizability improvement by an increase in percentage employed W\&H as the training set, resulting in an improvement of 7.4\%. We note that the W\&H dataset contains relatively the shortest average sentence length, and the HS samples show a direct HS style. These characteristics are also exhibited in experiments when the Founta dataset is employed as the training set, which shows the second highest generalizability improvement by percentage ($\up{}$5.1\%). Thus, the emotion knowledge supplied by the $GE_{ek}$ corpus is generally the better choice for improving cross-domain generalizability when a model is trained with short, explicit HS and is expected to generalize to HS that could be longer in length where the HS style might also be more implicit (e.g., Razavi, Kumar). For the case where the Offensive Reddit dataset is employed as the training set, we find that the HS samples in this dataset gear toward explicit sexism, which is not perceived in other datasets. In this case, the emotion knowledge supplied by the 28-class emotion corpus $GE_{go}$ helps to mitigate the semantic variance across contrasting HS topics more than the six-class $GE_{ek}$ corpus.

\subsection{Domain Adaptability of Base Models and Cross-Domain HS Generalizability}
\textbf{Observation 2:} \textit{The strength cross-generalizability enhancement is more pronounced with our emotion-enriched model when adopting a base model that is not adapted to the HS domain (e.g., BERT). However, adopting a base model that is adapted to the HS domain (e.g., fBERT) using our framework generally results in the highest cross-domain performance.}

To analyze the effect of the adopted base models' domain adaptability on cross-domain generalization, we visualize our framework's performance on cross-domain generalization using BERT-based and fBERT-based models eliminating the distinction of emotion corpora. We take the average performance of the cross-domain generalizability average of the variant of our model that uses the $GE_{go}$ corpus and the variant of our model that uses the $GE_{ek}$ for each base model. The average performance in binary F1 for the BERT-based and fBERT-based model are shown in Figure 3a. Figure 3b shows the percentage of cross-domain improvement relative to the baseline average.

From Figure 3, we note that adopting a non-HS domain-adapted model as the base model like BERT with our framework results in the greatest percentage of generalizability improvement in most cases. For the two cases where BERT-based emotion-enriched models show a relatively weaker generalizability uplift than fBERT-based models, we note that the training sets, Founta and W\&H datasets, are also the only two datasets that are sourced from Twitter.

Figure 3a supports that adopting a base model like fBERT that is adapted to the HS domain leads to higher performance despite the effect of adding emotion knowledge in uplifting cross-domain generalizability might not be as strong as adopting a non-HS domain adapted base model. For experiments where training sets are Kaggle, Kumar, Razavi, and W\&H, our emotion-enriched model that utilizes fBERT, which is pre-trained on a dataset that is in the HS domain, shows the best performance.

From Figure 3b, the most distinct improvement is in cross-domain settings with Kumar ($\up{}$7.6\%), Offensive Reddit ($\up{}$5.7\%), and W\&H ($\up{}$5.6\%) datasets as the training sets. As mentioned, the Offensive Reddit dataset exhibits a topical contrast to other datasets as its HS has a sexism focus. Hence, adopting a general base model with emotional knowledge helps to reduce the semantic variance across contrasting HS topics. We observe that the Kumar and W\&H datasets both exhibit relatively indirect styles of HS. This suggests that incorporating emotion knowledge helps to bridge the gap in allowing an HS model trained with a general non-HS domain adopted model on implicit HS to generalize on HS that are relatively more direct (e.g. Founta, Kaggle). 

\section{Conclusion}
In this work, we investigated cross-domain HS generalizability integrating emotion analysis. We presented a multitask HS generalizability framework that utilizes emotion knowledge to enhance cross-domain HS generalizability. We employed the 28-class GoEmotions corpus \citep{goemotions} and the traditional six-class Ekman corpus \citep{ekman1971universals} to examine their effects on improving cross-domain HS generalizability. We found that incorporating emotion knowledge using the Ekman corpus leads to more consistent generalizability performance. We also inspected the role of HS domain adaptiveness in base models on cross-domain HS generalizability and noted that the introduction of emotion knowledge has a relatively stronger strength in bridging the cross-domain generalization gap of pre-trained models that are not adapted to the HS domain. Results support that our emotion-enriched models outperform baselines in all cross-domain settings.

\section*{Limitations}
We acknowledge limitations in preserving the conceptual granularity exhibited in public HS datasets by adopting their varied categorical labels (i.e. toxic, abusive, sexism) in a binary form. Furthermore, the analyses presented in this work are based on the chosen datasets corresponding to their domain(s) only. Therefore, conclusions drawn from the limited quantity of datasets from restricted domains are not intended to be comprehensive. We also noted that more potential insights regarding HS generalizability may be drawn by comparing the results from evaluations against more state-of-the-art baselines with varied domain adaptiveness to specific aspects of HS (i.e. implicitness, sarcasm). This is left to future works.
 
\section*{Acknowledgements}
We thank all reviewers for their constructive feedbacks. This work is supported by in part by NSF 1946391 "RII Track-1: Data Analytics that are Robust and Trusted (DART): From Smart Curation to Socially Aware Decision Making."

\bibliography{anthology,custom}
\bibliographystyle{acl_natbib}

\end{document}

%% file: figtab/data.tex
\newcommand\tab[1][1cm]{\hspace*{#1}}

\begin{table*}[ht]
\centering
\begin{tabular}{lll}
\hline
\textbf{Dataset} & \textbf{Original Size} & \textbf{Domain}\\
\hline
Founta \citep{founta2018large} & 99,799 & Twitter \\
Kaggle \citep{kaggle} & 312,737 & Wikipedia \\
Kumar \citep{kumar} & 15,000 & Facebook \\
Offensive Reddit \citep{qian} &  5,020 & Reddit \\
Razavi \citep{razavi} & 1,525 & Natural Semantic Modules, Usenet \\
Waseem and Hovy \citep{waseem} & 16,907 & Twitter\\

\hline
GoEmotions \citep{goemotions} & 58,009 & Reddit\\
\hline

\end{tabular}
\caption{\label{data}
Datasets used in the cross-domain cross-dataset generalization evaluation. The GoEmotions \citep{goemotions} dataset supplies the auxiliary emotion knowledge in our multitask HS generalization framework.}
\end{table*}

%% file: figtab/emc.tex
\definecolor{lav}{rgb}{0.9, 0.9, 0.98}

\begin{table} [ht]
\centering
\begin{tabular}{ll}
\hline
\bm{$GE_{ek}$} & \bm{$GE_{go}$}\\
\hline
Anger & Anger, annoyance, disapproval  \\
Disgust & Disgust \\
Fear & Fear, nervousness  \\ 
Joy & Admiration, amusement, approval,   \\ 
& caring, desire, excitement,  \\
&  gratitude, joy, love,  \\
& optimism, pride, relief \\
 Sadness & Sadness, disappointment,  \\
& embarrassment, grief, remorse \\
Surprise & Surprise, realization,  \\  
& confusion, curiosity \\
\hline
\end{tabular}
\caption{Categorical emotion conversion between the version of the GoEmotions \citep{goemotions} dataset adopted with the Ekman $GE_{ek}$ \citep{ekman1971universals} corpus and its original $GE_{go}$ corpus.}
\label{tab:emc}
\end{table}

%% file: figtab/exp1.tex
\definecolor{lq}{rgb}{0.95, 0.95, 1}
\definecolor{lc}{rgb}{0.8, 0.9, 1}
\definecolor{lv}{rgb}{0.7, 0.8, 1}
\definecolor{ls}{rgb}{0.6, 0.7, 1}

\definecolor{iq}{rgb}{0.9, 0.75, 0.99}
\definecolor{ic}{rgb}{0.9, 0.8, 0.99}
\definecolor{iv}{rgb}{0.9, 0.85, 0.99}

\begin{table*} [ht]
\centering
\begin{tabular}{llllllll}
\hline
\textbf{Train/Test} & \textbf{Founta} & \textbf{Kaggle} & \textbf{Kumar} & \textbf{Off.Red.} & \textbf{Razavi} & \textbf{W\&H} & \textbf{CD Avg}\\
\hline
Founta  & 0.922 & 0.800 & 0.470 & 0.734 & 0.672 & 0.543 & 0.669\\
Founta + $GE_{go}$ & \cellcolor{iv} \textbf{0.926} & \cellcolor{lq} 0.809 & \cellcolor{ls} 0.556 & \cellcolor{lq} 0.739 & \cellcolor{lc} 0.686 & \cellcolor{iv} 0.553 & \cellcolor{lc} \textbf{0.697} \\
Founta + $GE_{ek}$  & \cellcolor{iv} 0.925 & \cellcolor{lc} 0.805 & \cellcolor{ls} 0.550 & \cellcolor{lq} 0.736 & \cellcolor{lc} 0.679 & \cellcolor{iv} 0.548 & \cellcolor{lc} 0.683 \\
\hline
Kaggle & 0.833 & 0.902 & 0.581 & 0.718 & 0.741 & 0.632 & 0.701\\
Kaggle + $GE_{go}$ & \cellcolor{lc} 0.845 & \cellcolor{iv} 0.911 & \cellcolor{lc} 0.592 & \cellcolor{lc} 0.725 & \cellcolor{lc} 0.767 & \cellcolor{lv} 0.677 & \cellcolor{lc} 0.721\\
Kaggle + $GE_{ek}$ & \cellcolor{lq} 0.839 & \cellcolor{iv} \textbf{0.916} & \cellcolor{lv} 0.616 & \cellcolor{lc} 0.742 & \cellcolor{lv} 0.777 & \cellcolor{ls} 0.701 & \cellcolor{lv} \textbf{0.735}\\
\hline
Kumar & 0.729 & 0.645 & 0.692 & 0.640 & 0.669 & 0.644 & 0.665\\
Kumar + $GE_{go}$ & \cellcolor{lv} 0.764 & \cellcolor{lv} 0.739 & \cellcolor{iv} 0.703 & \cellcolor{lv} 0.670 & \cellcolor{ls} 0.712 & \cellcolor{lc} 0.663 & \cellcolor{lv} 0.709 \\
Kumar + $GE_{ek}$ & \cellcolor{ls} 0.779 & \cellcolor{ls} 0.746 & \cellcolor{iv} \textbf{0.711} & \cellcolor{lv} 0.680 & \cellcolor{ls} 0.741 & \cellcolor{lc} 0.664 & \cellcolor{ls} \textbf{0.722} \\
\hline
Off. Red. & 0.671 & 0.638 & 0.530 & 0.931 & 0.627 & 0.618 & 0.617\\
Off. Red. + $GE_{go}$ & \cellcolor{ls} 0.727 & \cellcolor{ls} 0.694 & \cellcolor{lc} 0.558 & \cellcolor{iv} \textbf{0.932} & \cellcolor{lv} 0.657 & \cellcolor{lv} 0.650 & \cellcolor{lv} \textbf{0.657}\\
Off. Red. + $GE_{ek}$ & \cellcolor{lc} 0.688 & \cellcolor{lv} 0.678 & \cellcolor{lc} 0.556 & 0.931 & \cellcolor{lc} 0.651 & \cellcolor{lv} 0.659 & \cellcolor{lv} 0.646\\
\hline
Razavi & 0.798 & 0.829 & 0.566 & 0.767 & 0.866 & 0.631 & 0.718 \\
Razavi + $GE_{go}$ & \cellcolor{lv} 0.834 & \cellcolor{lq} 0.836 & \cellcolor{ls} 0.635 & \cellcolor{lq} 0.773 & \cellcolor{ic} \textbf{0.890} & \cellcolor{lc} 0.652 & \cellcolor{lc} 0.746\\
Razavi + $GE_{ek}$ & \cellcolor{lv} 0.844 & \cellcolor{lc} 0.855 & \cellcolor{ls} 0.663 & \cellcolor{lq} 0.770 & \cellcolor{iv} 0.871 & \cellcolor{lc} 0.653 & \cellcolor{lv} \textbf{0.757} \\
\hline
W\&H & 0.535 & 0.500 & 0.518 & 0.659 & 0.545 & 0.896 & 0.555 \\
W\&H + $GE_{go}$  & \cellcolor{iq} 0.597 & \cellcolor{lc} 0.529 & \cellcolor{ls} 0.570 & \cellcolor{lc} 0.675 & \cellcolor{lc} 0.557 & \cellcolor{iv} \textbf{0.899} & \cellcolor{lv} 0.583\\
W\&H + $GE_{ek}$  & \cellcolor{iq} 0.603 & \cellcolor{lv} 0.541 & \cellcolor{lv} 0.555 & \cellcolor{lc} 0.684 & \cellcolor{lc} 0.584 & \cellcolor{iv} 0.898 & \cellcolor{lv} \textbf{0.591} \\
\hline
\end{tabular}
\caption{\label{exp1} BERT-based in-domain and cross-domain HS generalization performance}
\end{table*}

%% file: figtab/exp2.tex
\definecolor{lq}{rgb}{0.9, 0.95, 1}
\definecolor{lc}{rgb}{0.8, 0.9, 1}
\definecolor{lv}{rgb}{0.7, 0.8, 1}
\definecolor{ls}{rgb}{0.6, 0.7, 1}

\definecolor{iq}{rgb}{0.9, 0.75, 0.99}
\definecolor{ic}{rgb}{0.9, 0.8, 0.99}
\definecolor{iv}{rgb}{0.9, 0.85, 0.99}

\begin{table*}[ht]
\centering
\begin{tabular}{llllllll}
\hline
\textbf{Train/Test} & \textbf{Founta} & \textbf{Kaggle} & \textbf{Kumar} & \textbf{Off.Red.} & \textbf{Razavi} & \textbf{W\&H} & \textbf{CD Avg}\\
\hline
Founta  & 0.929 & 0.7961 & 0.377 & 0.743 & 0.600 & 0.526 & 0.629 \\
Founta + $GE_{go}$ & \cellcolor{iv} \textbf{0.930} & \cellcolor{lq} 0.805 & \cellcolor{lc}  0.397 & \cellcolor{lc}  0.753 & \cellcolor{lq} 0.609 & \cellcolor{iq} 0.560 & \cellcolor{lc} 0.625 \\
Founta + $GE_{ek}$  & 0.929 & \cellcolor{lv} 0.841 & \cellcolor{lv} 0.411 & \cellcolor{lc} 0.751 & \cellcolor{ls} 0.662 &\cellcolor{iq} 0.576 & \cellcolor{lv} \textbf{0.666} \\
\hline
Kaggle & 0.848 & 0.922 & 0.596 & 0.745 & 0.765 & 0.756 & 0.742\\
Kaggle + $GE_{go}$ & \cellcolor{lc} 0.854 & \cellcolor{iv} 0.925 & \cellcolor{lc} 0.609 & \cellcolor{lc} 0.754 &\cellcolor{lc} 0.775 & \cellcolor{lc} 0.769 & \cellcolor{lc} 0.752\\
Kaggle + $GE_{ek}$ & \cellcolor{lc} 0.861 & \cellcolor{iv} \textbf{0.926} & \cellcolor{lc} 0.618 & \cellcolor{lc} 0.757 & \cellcolor{lc} 0.785 & \cellcolor{lc} 0.783 & \cellcolor{lc} \textbf{0.761}\\
\hline
Kumar & 0.848 & 0.596 & 0.715 & 0.596 & 0.765 & 0.756 & 0.712\\
Kumar + $GE_{go}$ & \cellcolor{lc} 0.868 & \cellcolor{lc} 0.619 & \cellcolor{iv} 0.721 & \cellcolor{lc} 0.611 & \cellcolor{lv} 0.797 & \cellcolor{lv} 0.787 & \cellcolor{lc} \textbf{0.736}\\
Kumar + $GE_{ek}$ & \cellcolor{lc} 0.857 & \cellcolor{lq} 0.602 & \cellcolor{iv} \textbf{0.733} & \cellcolor{lq} 0.600 & \cellcolor{lv} 0.797 & \cellcolor{lv} 0.787 & \cellcolor{lc} 0.729\\
\hline
Off. Red. & 0.644 & 0.631 & 0.537 & 0.936 & 0.621 & 0.659 & 0.618 \\
Off. Red. + $GE_{go}$ & \cellcolor{lv} 0.682 & \cellcolor{ls} 0.688 & \cellcolor{lc} 0.550 & 0.933 & \cellcolor{lv} 0.665 & \cellcolor{lv} 0.683 & \cellcolor{lv} \textbf{0.653} \\
Off. Red. + $GE_{ek}$ & \cellcolor{lc} 0.652 & \cellcolor{lc} 0.645 & \cellcolor{lc} 0.553 & \cellcolor{iv} \textbf{0.938} & \cellcolor{lv} 0.665 & \cellcolor{lv} 0.700 & \cellcolor{lv} 0.643 \\
\hline
Razavi & 0.845 & 0.871 & 0.642 & 0.768 & 0.881 & 0.790 & 0.783 \\
Razavi + $GE_{go}$ & \cellcolor{lv} 0.877 & \cellcolor{lq} 0.879 & \cellcolor{lq} 0.643 & \cellcolor{lq} 0.771 & 0.878 & \cellcolor{lq} 0.791 & \cellcolor{lq} \textbf{0.792} \\
Razavi + $GE_{ek}$ & \cellcolor{lc} 0.859 & \cellcolor{lq} 0.872 & \cellcolor{lq} 0.644 & \cellcolor{lq} 0.769 & \cellcolor{iv} \textbf{0.885} & \cellcolor{lq} 0.795 & \cellcolor{lq} 0.788 \\
\hline
W\&H & 0.607 & 0.605 & 0.505 & 0.729 & 0.526 & 0.894 & 0.526\\
W\&H + $GE_{go}$  & \cellcolor{iq} 0.663 & \cellcolor{lc} 0.634 & \cellcolor{ls} 0.574 & \cellcolor{lc} 0.742 & \cellcolor{ls} 0.607 & \cellcolor{iv} 0.896 & \cellcolor{ls} 0.638 \\
W\&H + $GE_{ek}$  & \cellcolor{ic} 0.630 & \cellcolor{lc} 0.630 & \cellcolor{ls} 0.577 & \cellcolor{lc} 0.753 & \cellcolor{ls} 0.603 & \cellcolor{iv} \textbf{0.903} & \cellcolor{ls} \textbf{0.640}\\
\hline
\end{tabular}
\caption{\label{exp2} fBERT-based in-domain and cross-domain HS generalization performance}
\end{table*}